\documentclass[lettersize,journal]{IEEEtran}
\usepackage{amsmath,amsfonts}
\usepackage{algorithmic}
\usepackage{algorithm}
\usepackage{array}
\usepackage[caption=false,font=normalsize,labelfont=sf,textfont=sf]{subfig}
\usepackage{textcomp}
\usepackage{stfloats}
\usepackage{url}
\usepackage{verbatim}
\usepackage{graphicx}
\graphicspath{{figs/}}
\usepackage{cite}
\usepackage{siunitx}
\usepackage{booktabs}
\usepackage{multirow}
\usepackage{mathtools}
\usepackage{rotating}
\usepackage{bbm} 
\newcommand{\norm}[1]{\left\lVert#1\right\rVert}

\usepackage{xspace}
\makeatletter
\DeclareRobustCommand\onedot{\futurelet\@let@token\@onedot}
\def\@onedot{\ifx\@let@token.\else.\null\fi\xspace}
\def\aka{{\it a.k.a}\onedot} 
\def\eg{{\it e.g}\onedot} 
\def\ie{{\it i.e}\onedot} 
 
\def\etc{{\it etc}\onedot} 
 
\def\etal{{\it et al}\onedot}
\makeatother

\begin{document}

\title{Efficient Key-Based Adversarial Defense for ImageNet by Using Pre-trained Model}

\author{AprilPyone MaungMaung$^1$, Isao Echizen$^{1,2}$, and Hitoshi Kiya$^{3}$ \\
\small{$^1$National Institute of Informatics, Tokyo, Japan\qquad $^2$University of Tokyo, Tokyo, Japan} \qquad
\small{$^3$Tokyo Metropolitan University, Tokyo, Japan}
}

\maketitle

\begin{abstract}
In this paper, we propose key-based defense model proliferation by leveraging pre-trained models and utilizing recent efficient fine-tuning techniques on ImageNet-1k classification.
First, we stress that deploying key-based models on edge devices is feasible with the latest model deployment advancements, such as Apple CoreML, although the mainstream enterprise edge artificial intelligence (Edge AI) has been focused on the Cloud.
Then, we point out that the previous key-based defense on on-device image classification is impractical for two reasons: (1) training many classifiers from scratch is not feasible, and (2) key-based defenses still need to be thoroughly tested on large datasets like ImageNet.
To this end, we propose to leverage pre-trained models and utilize efficient fine-tuning techniques to proliferate key-based models even on limited compute resources.
Experiments were carried out on the ImageNet-1k dataset using adaptive and non-adaptive attacks.
The results show that our proposed fine-tuned key-based models achieve a superior classification accuracy (more than $10\%$ increase) compared to the previous key-based models on classifying clean and adversarial examples.
\end{abstract}

\section{Introduction}
Deep learning has brought breakthroughs in many applications~\cite{lecun2015deep}.
Some notable examples are visual recognition~\cite{he2016deep}, natural language processing~\cite{vaswani2017attention}, and speech recognition~\cite{graves2013speech}.
Despite the remarkable performance, machine learning (ML) generally, including deep learning, is vulnerable to various attacks.
Notably, many ML algorithms, including deep neural networks, are sensitive to carefully perturbed data points known as adversarial examples intentionally designed to cause ML models make mistakes~\cite{szegedy2013intriguing,biggio2013evasion,goodfellow2015explaining}.
In many cases, perturbation added to make adversarial examples is often imperceptible to humans but still causes ML models to make erroneous predictions with high confidence.
Previous works have proven that adversarial examples can be applied to real-world scenarios~\cite{kurakin2018adversarial,athalye2018synthesizing}.
Adversarial examples can potentially be dangerous, especially for autonomous vehicles~\cite{papernot2017practical,eykholt2018robust}.

As adversarial examples are an obvious threat, researchers have proposed numerous methods to defend against adversarial examples in the literature~\cite{yuan2019adversarial}.
However, most defense methods either reduce the classification or are completely broken by adaptive attacks~\cite{athalye2018obfuscated,tramer2020adaptive}.
Therefore, defending against adversarial examples is still challenging and remains an open problem.
Inspired by cryptography, a new line of research on adversarial defense has focused on using secret keys so that defenders have some information advantage over attackers~\cite{taran2018bridging,aprilpyone2021block,kiya2022overview,maung2020encryption,iijima2023enhanced}.
Key-based defenses follow Kerckhoffs's second cryptographic principle, which states that a system should not require secrecy even if it is exposed to attackers, but the key should be secret~\cite{kerckhoffs1883cryptographic}.
By keeping a secret key, key-based defenses make adversarial attacks ineffective.
The idea of making adversarial attacks expensive or ideally intractable is further supported on a theoretical basis that adversarially robust machine learning could leverage computational hardness, as in cryptography~\cite{garg2020adversarially}.
To further harden key-based defenses, researchers have also proposed to use implicit neural representation~\cite{rusu2022hindering} and ensembles of key-based defenses~\cite{taran2020machine,maungmaung2021ensemble}.

\begin{figure}
\centerline{\includegraphics[width=\linewidth]{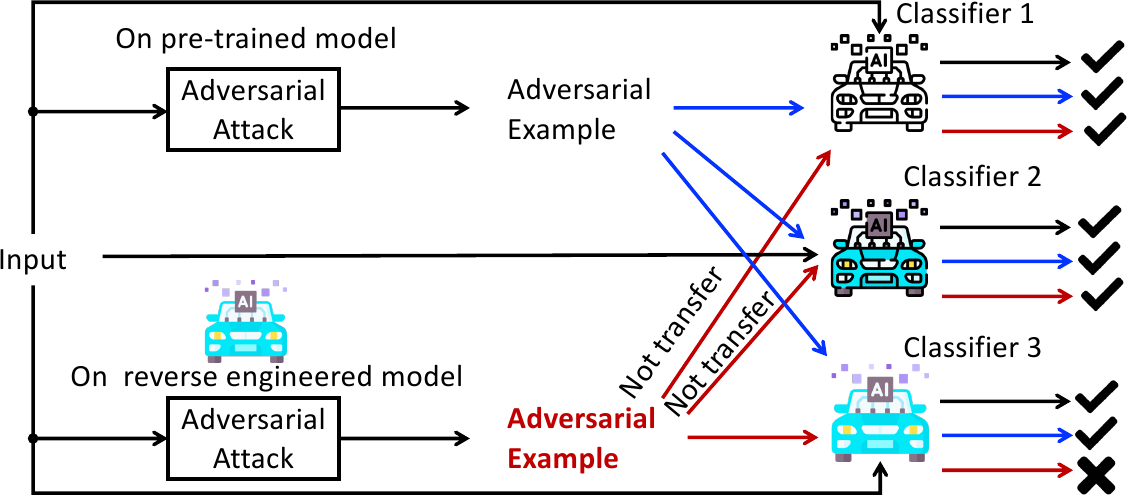}}
\caption{Scenario of on-device image classification under adversarial settings.\label{fig:scenario}}
\end{figure}

One of the advantages of key-based defenses is that a classifier has its own key.
This feature is handy as adversarial examples are transferable within models with the same architecture~\cite{su2018robustness} or different ones~\cite{liu2017delving}.
Therefore, in this paper, we consider an on-device image classification scenario (Fig.~\ref{fig:scenario}) where each classifier has its own key.
If an attacker successfully reverse engineers a classifier (\eg, classifier 3 in Fig.~\ref{fig:scenario}), adversarial examples generated for the compromised model cannot transfer to other models.

The use case of the one-key-one-model approach is relevant and realistic, especially nowadays, because even large models can be deployed on CPUs~\cite{shen2023efficient} and on edge devices~\cite{coremlstable}.
For example, Apple Silicon devices can run a large image generative model like Stable Diffusion~\cite{coremlstable}.
In addition, even large language models (LLMs) can be deployed on edge devices by using C/C++ implementation of models with the ggml\footnote{\scriptsize \url{https://github.com/ggerganov/ggml}} library.
For image classification, vit.cpp\footnote{\scriptsize \url{https://github.com/staghado/vit.cpp}} (C++ inference engine for vision transformer models) is available for edge devices.
Although deep learning models are widely deployed on edge devices and will be deployed more and more in the near future, the one-key-one-model approach is underexplored.

Therefore, we focus on the one-key-one-model scenario in this paper.
First, we point out that key-based defenses seem promising but are not practical for two reasons: (1) training many classifiers from scratch is not feasible, and (2) key-based defenses have not yet been thoroughly tested on large datasets like ImageNet.
Then, we build upon the idea of key-based defense and propose to leverage pre-trained models and use the latest fine-tuning techniques to train many defended models efficiently.
In experiments, our key-based models are efficiently trained and achieve a higher classification accuracy for both clean and p-norm bounded adversarial examples for ImageNet-1k classification compared to state-of-the-art methods.
However, key-based models have information advantage over attackers for having a secret key.
We make the following contributions in this paper.
\begin{itemize}
\item We stress that the one-key-one-model approach is relevant nowadays as more and more models are being deployed on edge devices, and key-based defenses have the potential to combat adversarial examples that are often transferred from one model to another.
\item We propose to leverage pre-trained models and use the latest fine-tuning techniques to train key-based models efficiently for the first time.
\item We conduct experiments for ImageNet-1k classification and evaluate key-based models using adaptive and non-adaptive attacks.
\end{itemize}

The rest of this paper is structured as follows.
Section~\ref{sec:related} presents related work on adversarial examples, defenses, pre-trained models, and a recent fine-tuning technique, LoRA.\@
Section~\ref{sec:proposed} puts forward the proposed defense.
Experiments on various attacks including adaptive ones are presented in Section~\ref{sec:experiments}, and Section~\ref{sec:discussion} presents discussion on the proposed defense.
Then, Section~\ref{sec:conclusion} concludes this paper.

\section{Related Work\label{sec:related}}
\subsection{Adversarial Examples}
Adversarial examples~\cite{szegedy2013intriguing,biggio2013evasion,goodfellow2015explaining} are intentionally perturbed inputs to machine learning models that cause the model to make erroneous predictions~\cite{goodfellowblog}.
There are two kinds of adversarial examples based on how they are generated: perturbation-based and unrestricted adversarial examples.
Perturbation-based adversarial examples are generally p-norm bounded, and different matrix norms such as $\ell_{\infty}$~\cite{madry2018towards}, $\ell_2$~\cite{moosavi2016deepfool}, $\ell_1$~\cite{chen2018ead}, and $\ell_0$~\cite{papernot2016limitations} are used to restrict the perturbation.
Beyond norm-bounded perturbation, adversarial examples can also be found in an unrestricted way~\cite{brown2018unrestricted} by applying spatial transformation~\cite{engstrom2019exploring} or generative models~\cite{song2018constructing}.

Adversarial examples can generalize to real-world applications and have the potential to be dangerous.
Kurakin~\etal showed that an adversarial example can be photographed with a smartphone, and the taken picture can still fool a model~\cite{kurakin2018adversarial}.
Researchers have also demonstrated that it is possible to construct 3D adversarial objects~\cite{athalye2018synthesizing}.
The threat of adversarial examples is especially alarming for autonomous vehicles.
Attackers could create stickers or paint to design adversarial stop signs to cause accidents~\cite{papernot2017practical,eykholt2018robust}.
In addition, adversarial examples can be deployed in many different ways to fool facial recognition or objection detection systems, such as adversarial t-shirts~\cite{xu2020adversarial}, adversarial hats~\cite{komkov2021advhat}, adversarial glasses~\cite{sharif2016accessorize}, adversarial make-up~\cite{guetta2021dodging}, \etc.
Generating real-world adversarial examples is not limited to small or imperceptible changes to the input.
Adversarial examples can also be crafted by placing a small visible image-independent patch~\cite{brown2017adversarial}.

In this paper, we still deploy p-norm bounded adversarial examples since they are well-defined to evaluate the proposed defense.
However, we consider a realistic attacking scenario.

\subsection{Adversarial Defenses}
There are two distinct strategies in designing adversarial defenses.

\vspace{2mm}\noindent{\bf (1)}
Classifiers are designed in such a way that they are robust against all adversarial examples in a specific adversarial space either empirically (\ie, adversarial training) or in a certified way (\ie, certified defenses).
Current empirically robust classifiers utilize adversarial training, which includes adversarial examples in a training set.
Madry~\etal approach adversarial training as a robust optimization problem and utilize projected gradient descent (PGD) adversary under $\ell_\infty$-norm to approximate the worst inputs possible (\ie, adversarial examples)~\cite{madry2018towards}.
As PGD is iterative, the cost of computing PGD-based adversarial examples is expensive.
Much progress has been made in reducing the computation cost of adversarial training, such as free adversarial training~\cite{shafahi2019adversarial}, fast adversarial training~\cite{wong2020fast}, and single-step adversarial training~\cite{de2022make}.
However, adversarially trained models (with $\ell_\infty$ norm-bounded perturbation) can still be attacked by $\ell_1$ norm-bounded adversarial examples~\cite{sharma2017attacking}.

Another approach is to use formal verification methods in such a way that no adversarial examples exist within some bounds~\cite{wong2018provable,raghunathan2018certified,cohen2019certified,hein2017formal}.
Ideally, these defenses are preferred for achieving certain guarantees.
Although certified defenses are attractive, they can be bypassed by generative perturbation~\cite{poursaeed2018generative} or parametric perturbation (outside of pixel norm ball)~\cite{liu2018beyond}.

\vspace{2mm}\noindent{\bf (2)}
Another primary strategy for designing adversarial defenses is that input data to classifiers are pre-processed in such a way that adversarial examples are ineffective (\ie, input transformation defenses, key-based defenses).
The idea is to find a defensive transform to reduce the impact of adversarial noise or make adversarial attacks ineffective (\ie, computing adversarial noise is either expensive or intractable).
The works in this direction use various transformation methods, such as thermometer encoding~\cite{buckman2018thermometer}, diverse image processing techniques~\cite{guo2018countering,xie2018mitigating}, denoising strategies~\cite{liao2018defense,niu2020limitations}, GAN-based transformation~\cite{song2018pixeldefend}, and so on.\
Although these input transformation-based defenses provided high accuracy at first, they can be attacked by adaptive attacks such as~\cite{athalye2018obfuscated,tramer2020adaptive}.
Unlike input transformation-based defenses, key-based defenses have an information advantage over attackers.
Such key-based defenses include~\cite{taran2018bridging,aprilpyone2021block,rusu2022hindering}.

Inspired by cryptography, the main idea of the key-based defense is to embed a secret key into the model structure with minimal impact on model performance.
Assuming the key stays secret, an attacker will not obtain any useful information on the model, which will render adversarial attacks ineffective.
Generally, key-based defenses hide a model’s decision from attackers by means of training the model with encrypted images.
In this paper, we adopt block-wise pixel shuffling from the key-based defense~\cite{aprilpyone2021block}.

\subsection{Pre-trained Models}
Pre-training is a favorable paradigm for many computer vision tasks because training a large-scale model is a non-trivial task and requires a significant amount of resources.
Therefore, it is not feasible for many users to train a large-scale model from scratch.
Besides the training cost, pre-training also improves generalization for downstream tasks~\cite{yosinski2014transferable}.
Recent works on the pre-training ImageNet-21k dataset~\cite{deng2009imagenet} (approximately 14 million images with about 21,000 distinct object categories) show superior performance on ImageNet-1k classification~\cite{steiner2022how,ridnik2021imagenet}.

Moreover, current frontiers of AI applications such as ChatGPT are driven by pre-trained models (\aka foundation models).
The term foundation model was coined by researchers from the Stanford Institute for Human-Centered Artificial Intelligence (HAI).
A foundation model is a large-scale model trained with a vast amount of data (generally using self-supervised learning), and it can be adapted to a wide range of downstream applications such as generating text and images, understanding natural language, \etc~\cite{bommasani2021opportunities}.

We are motivated by the success of pre-trained models, and we utilize an ImageNet-21k pre-trained vision transformer (ViT)~\cite{dosovitskiy2020image} in this paper.

\subsection{LoRA}
LoRA, which stands for low-rank adaptation of large language models is an efficient fine-tuning technique initially introduced to fine-tune large language models~\cite{hu2022lora}.
Instead of fine-tuning the whole model, LoRA first freezes all the pretrained model weights and then injects trainable rank decomposition matrices into each layer of the transformer block.
Therefore, the number of trainable parameters of a model is significantly reduced.
Compared to full fine-tuning, LoRA maintains a competitive performance without increasing any additional inference latency.
In this paper, we utilize LoRA to fine-tune a pre-trained ViT model to be a key-based defended ViT model.

\section{Proposed Defense\label{sec:proposed}}
\subsection{Requirements}
We aim to fulfill the following requirements in the proposed defense.
\begin{itemize}
  \item Leverage pre-trained models and fine-tune them efficiently so that training a key-based model on an ImageNet scale is practical for the majority of users.
  \item Achieve high classification accuracy for both clean and adversarial examples.
  \item Associate one key to one model only so that a successful attack on one model does not transfer to another model with another key.
\end{itemize}

\subsection{Threat Model\label{sec:threat}}
A threat model includes a set of assumptions, such as an adversary’s goals, knowledge, and capabilities~\cite{carlini2019evaluating}.

As we focus on image classifiers, the adversary's goal is to change the predicted class from a true class either in a targeted or untargeted way.
We deploy the AutoAttack (AA) strategy, which is a suite of both targeted and untargeted attacks~\cite{croce2020reliable} in this paper.

The adversary's knowledge can be classified as white-box, black-box, or gray-box.
In white-box settings, the adversary has complete knowledge of the model, its parameters, training data, and the inner workings of the defense mechanism.
In contrast, the black-blox adversary has no knowledge about the model.
However, in many cases, the adversary knows something in between white-box and black-box, which is referred to as a gray-box scenario.
Since the gray-box scenario is more realistic, we consider that the adversary knows the model architecture and has access to pre-trained models and training data.
In addition, we assume the gray-box adversary also knows the mechanism of the key-based defense, but not the secret key.

We consider a p-norm bounded threat model for its well-defined nature.
Therefore, the adversary can add small perturbation $\delta$ under some budget $\epsilon$ (\ie, $\norm{\delta}_p \le \epsilon$), where $\epsilon > 0$.

\begin{figure}
\centerline{\includegraphics[width=\linewidth]{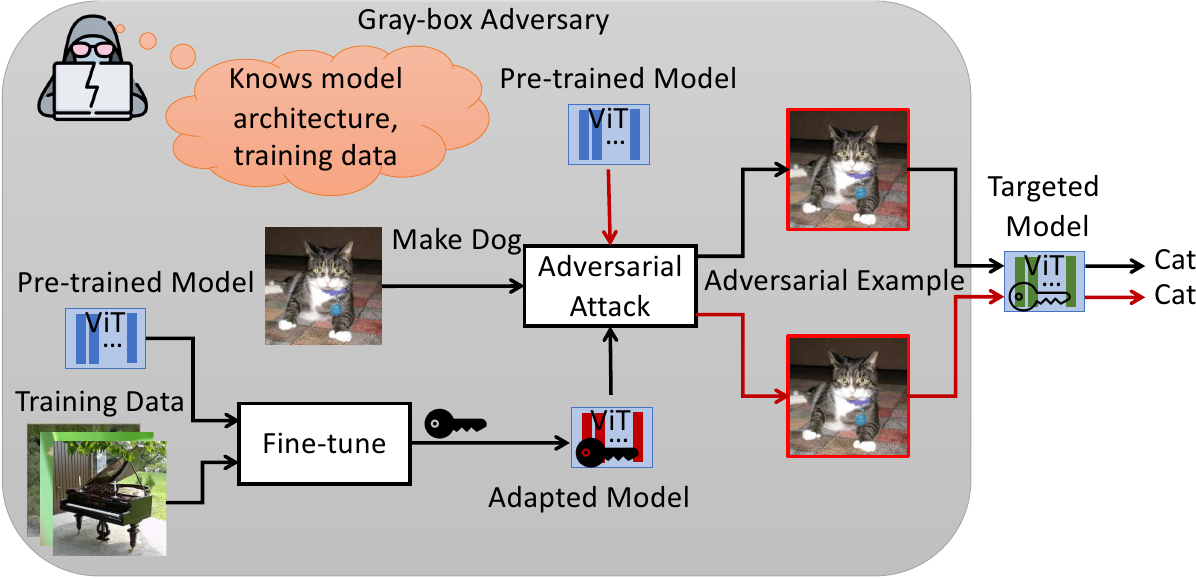}}
\caption{Assumed threat model.
The adversary carries out attacks on both pre-trained model and adapted fine-tuned model.\label{fig:threat}}
\end{figure}

With the above assumptions, Fig.~\ref{fig:threat} shows the considered gray-box adversary that performs attacks on a pre-trained model or an adapted fine-tuned model with an assumed key to target a defended model.

\subsection{Overview}
\begin{figure}
\centerline{\includegraphics[width=\linewidth]{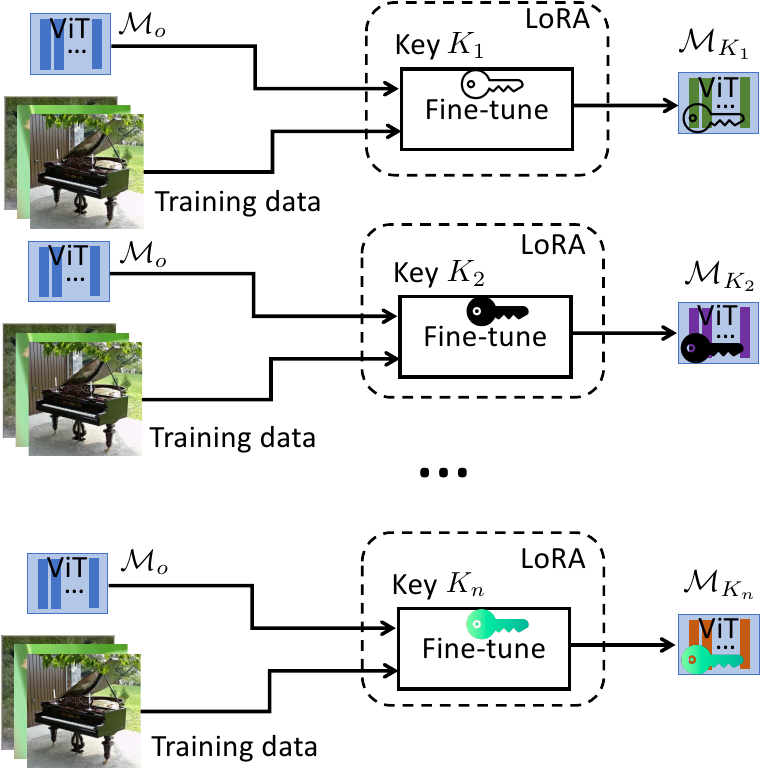}}
\caption{Proposed defense.
A pre-trained model is fine-tuned with many keys to produce many defended models.\label{fig:overview}}
\end{figure}

We consider an on-device image classification scenario.
Such a scenario is practical because, nowadays, devices such as autonomous vehicles or smartphones are equipped with image classifiers.
The basic idea of the proposed defense is to personalize a pre-trained model with a secret key efficiently so that a classifier has its own decision-making process based on a key.
By leveraging the previous works, key-based defense~\cite{aprilpyone2021block}, training improvements~\cite{steiner2022how}, and efficient fine-tuning~\cite{hu2022lora}, we propose to efficiently fine-tune a pre-trained model to many defended models with many different keys, as shown in Fig.~\ref{fig:overview} in this paper.
Consequently, defended models can be potentially deployed on devices such that one defended model is associated with one key only.

\subsection{Training}
Given a dataset $\mathcal{D}$ with pairs of examples (images and corresponding labels), $\{(x, y) \;|\; x \in \mathcal{X},\; y \in \mathcal{Y}\}$,
a key-based defense maps the input space $\mathcal{X}$ to an encrypted space $\mathcal{H}$ by using some transformation $t$ controlled by a secret key $K$ (\ie, $x \mapsto t(x, K)$).
A classifier $f: \mathcal{H} \rightarrow \mathcal{Y}$ is trained by using encrypted images $t(x, K) \in \mathcal{H}$.
For simplicity, we define such a defended model with key $K$ as
\begin{equation}
  \mathcal{M}_K(\cdot) = f(t(\cdot, K)),
\end{equation}
where $t(\cdot, K)$ is a key-based transformation and $f$ is a deep neural network-based image classifier such as ViT~\cite{dosovitskiy2020image}.
In this paper, we adopt block-wise pixel shuffling from~\cite{aprilpyone2021block} as $t(\cdot, K)$ and the detailed procedure is as follows.

\begin{enumerate}
  \item Divide a three-channel (RGB) color image, $x$ with $w \times h$ into non-overlapping blocks, each with $P \times P$ such that $\{B_1, \ldots, B_i, \ldots, B_{(\frac{h}{P} \times \frac{w}{P})}\}$.
    \item Generate a random permutation vector, $v$ with key $K$, such that \\ $(v_1, \ldots, v_k, \ldots, v_{k'}, \ldots, v_{3P^2})$, where $v_k \neq v_{k'}$ if $k \neq k'$.
    \item For each block $B_i$, \\
      flatten three-channel block of pixels into a vector, $b_i$ such that $b_i = (b_i(1), \ldots, b_i({3P^2}))$,\\
      permute pixels in $b_i$ with $v$ such that
      \begin{equation}
        b'_i(k) = b_i(v_k), k \in \{1, \ldots, 3P^2\},
      \end{equation}
      and reshape the permuted  vector $b'_i$ back into the three-channel block $B'_i$.
    \item Integrate all permuted blocks, $B'_1$ to $B'_{(\frac{h}{P} \times \frac{w}{P})}$ to form a three-channel pixel shuffled image, $x'$.
\end{enumerate}

A key-based defended model $\mathcal{M}_{K}$ is obtained by fine-tuning a pre-trained model $\mathcal{M}_o$ (which is trained by using plain images) with transformed images by key $K$.
By leveraging efficient fine-tuning techniques such as LoRA~\cite{hu2022lora}, many defended models ($\mathcal{M}_{K_1}, \mathcal{M}_{K_2}, \ldots, \mathcal{M}_{K_n}$) can be efficiently fine-tuned given keys ($K_1, K_2, \ldots, K_n$) as shown in Fig.~\ref{fig:overview}.
We are the first to consider LoRA in the key-based defense.

The previous work~\cite{maung2023hindering} showed that a key-based defended model can be obtained by fine-tuning only the patch embedding layer and the classifier head of an isotropic convolutional neural network.
However, the performance accuracy dropped.
We further extend the previous work~\cite{maung2023hindering} by applying LoRA to both the patch embedding layer and the transformer block to improve the performance accuracy in this paper.

\subsection{Inference}
During inference, a defended model $\mathcal{M}_{K}$ first transforms test images with key $K$ prior to image classification procedure by a deep neural network backbone such as ViT.
A predicted class label $\hat{y}$ is obtained from a defended classifier $\mathcal{M}_{K}$ as
\begin{equation}
  \hat{y} = \text{arg\,max}_i {\mathcal{M}_{K}(x)}_{i} = \text{arg\,max}_i {f(t(x, K))}_i.
\end{equation}

\subsection{Evaluation Metric\label{sec:metric}}
To evaluate the proposed defense, we calculate accuracy for classifying both clean and adversarial examples. The accuracy is computed as
\begin{equation}
 \text{Accuracy} = \left\{ \begin{array}{ll}
     \frac{1}{N}\sum_{i=1}^{N} \mathbbm{1} (\mathcal{M}_K(x_i) = y_i) & (\text{clean})\\
     \frac{1}{N}\sum_{i=1}^{N} \mathbbm{1} (\mathcal{M}_K(x_i + \delta{i}) = y_i) & (\text{attacked}),
 \end{array}
 \right.
\end{equation}
where $\mathcal{M}_K$ is a defended classifier with key $K$, $N$ is the number of test images, $\mathbbm{1}(\text{condition})$ is one if condition is true, otherwise zero, $\{x_i, y_i\}$ is a test image ($x_i$) with its corresponding label ($y_i$), and $\delta_i$ is its respective adversarial noise depending on a specific attack.

\section{Experiments\label{sec:experiments}}
\subsection{Setup}
\noindent{\bf Datasets.}
We carried out ImageNet-1k classification experiments for the proposed defense.
We utilized the ImageNet-1k dataset (with 1000 classes) consisting of 1.28 million color images for training and 50,000 color images for validation~\cite{ILSVRC15}.
ImageNet-1k was introduced for the ILSVRC2012 visual recognition challenge and is regarded as one of the main datasets for image classification research.
It is a subset of the ImageNet-21k dataset, which consists of approximately 14 million images with about 21,000 classes~\cite{deng2009imagenet}.
We resized all images to a dimension of $224 \times 224$.

\vspace{2mm}\noindent{\bf Models.}
We utilized pre-trained ViT base models with a patch size value of $16$ from~\cite{steiner2022how} and~\cite{dosovitskiy2020image} that were trained on ImageNet-21k.
To implement the proposed defense, we modified pytorch image models\footnote{\scriptsize \url{https://github.com/huggingface/pytorch-image-models}}, and LoRA fine-tuning\footnote{\scriptsize \url{https://github.com/huggingface/peft}}.
We followed the training settings from~\cite{steiner2022how} to fine-tune the proposed defended models.
For LoRA, we used a low-rank dimension value $r = 16$ and a value of scaling factor $16$.

\vspace{2mm}\noindent{\bf Attacks.}
For evaluation, we deployed the AutoAttack (AA) strategy, which is an ensemble of strong, diverse attacks consisting of both targeted and untargeted attacks~\cite{croce2020reliable}.
We used AA's ``standard'' version with a perturbation budget value of $4/225$ under $\ell_\infty$ norm and a value of $0.5$ under $\ell_2$ norm for all attacks.

\subsection{Classification Accuracy}
We fine-tuned key-based models with or without LoRA and performed ImageNet-1k classification on both clean images and adversarial examples.
Table~\ref{tab:results} summarizes results of both clean and robust accuracy compared to state-of-the-art key-based methods~\cite{maung2023hindering,aprilpyone2021block} and adversarial training (AT) methods~\cite{liu2023comprehensive,singh2023revisiting}.
In the table, (--) denotes ``there are no reported results'' or ``not applicable,'' and both clean and robust accuracies were calculated on the validation set of ImageNet-1k (50,000) images as described in Section~\ref{sec:metric}.
Although the un-defended model (plain) achieved the highest accuracy, it was most vulnerable to all attacks.
In terms of clean accuracy, our best model (fully fine-tuned) with block size ($P = 4$) achieved $83.40\%$, which is about $1.7\%$ lower compared to the plain model.

\begin{figure}
\centerline{\includegraphics[width=\linewidth]{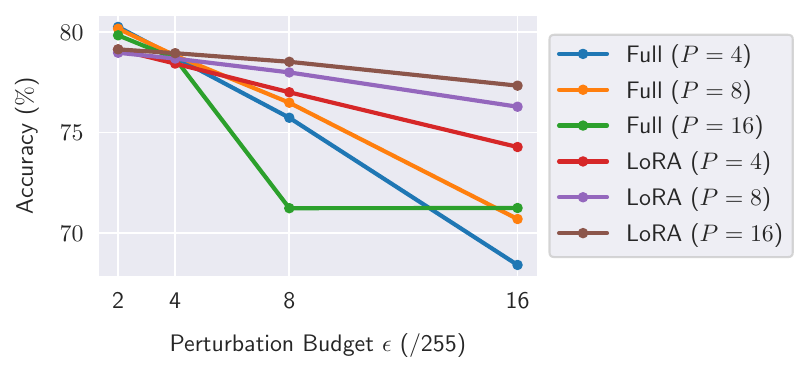}}
\caption{Accuracy ($\%$) of key-based fine-tuned models under AA ($\ell_\infty)$ adversary with various perturbation budget $\epsilon$. The accuracy was calculated over 50,000 images (whole ImageNet-1k validation set).\label{fig:linf}}
\end{figure}

\vspace{2mm}\noindent{\bf Non-Adaptive Attacks.}
Since the proposed defense is built on top of the pre-trained model, it is natural to generate adversarial examples directly on the pre-trained model.
We refer to this scenario as a non-adaptive attack.
The results from Table~\ref{tab:results} show that adversarial examples generated on the pre-trained model were not effective on either fully fine-tuned or LoRA models.
In addition, we also carried out non-adaptive attacks with $\ell_\infty$ adversary with different perturbation budget $\epsilon$.
As shown in Fig.~\ref{fig:linf}, as we increased $\epsilon$, the accuracy gradually dropped as expected.

\vspace{2mm}\noindent{\bf Adaptive Attacks.}
We assume the gray-box adversary as described in Section~\ref{sec:threat}.
Thus, the adversary has knowledge of the key-based defense mechanism except the key.
The attacker may fine-tune the pre-trained model with a guessed key and prepare a similar substitute model as the targeted key-based model.
By using the substitute key-based model, the adversary may generate adversarial examples.
We refer to this scenario as an adaptive attack.
From the results (Table~\ref{tab:results}), our fully fine-tuned models were robust against such attacks.
Interestingly, adaptive attacks were even worse than non-adaptive attacks on fully fine-tuned models.
However, adaptive attacks were successful on LoRA fine-tuned models, especially for the models with $P = 8$ and $16$.
We suspect that the LoRA applied patch embedding layer in ViT-B/16 does not fully capture the key-based transformation.
We shall further investigate the reason and improve the robustness of LoRA fine-tuned models in our future work.

\vspace{2mm}\noindent{\bf Comparison with State-of-the-Art Methods.}
We compare the proposed fine-tuned models with the previous key-based models~\cite{maung2023hindering,aprilpyone2021block}, and the top 3 adversarially trained (AT) models from RobustBench~\cite{croce2020robustbench} in terms of clean and robust accuracy.
Note that it is not fair to compare key-based models directly with AT models because key-based models have an information advantage (secret key) over attackers, while AT models do not.
The previous key-based methods utilized different model architectures, ConvMixer-768/32 and ResNet50, which are much smaller compared to ViT-B/16.
The ConvMixer one used partial fine-tuning (patch embeddings and classifier head), and the ResNet50 one employed full-finetuning.
In contrast, we utilized a larger model, ViT-B/16, with the latest fine-tuning techniques for efficient training in this paper.
Our models achieved a superior performance (more than a $10\%$ increase) on both clean and robust accuracy compared to the previous key-based models and AT models.

\begin{table*}
  \caption{Clean and Robust Accuracy ($\%$) of Proposed Key-Based Models and State-of-the-Art Models\label{tab:results}}
  \centering
  \resizebox{\linewidth}{!}{%
  \begin{tabular}{lcccccccc} 
    \toprule
    \multirow{2}{*}{Model} & \multirow{2}{*}{Defense} & {Block Size} & \multirow{2}{*}{Fine-tune} & \multirow{2}{*}{Clean} & \multicolumn{2}{c}{Robust $\ell_\infty$ ($\epsilon = 4/255$)} & \multicolumn{2}{c}{Robust $\ell_2$ ($\epsilon = 0.5$)}\\
                           & & ($P$) & & & (Non-Adaptive)  & (Adaptive) & (Non-Adaptive)  & (Adaptive)\\
    \midrule
    ViT-B/16 & {Plain} & {--} & {No} & 85.10 & 0.0 & {--} & 0.32 & {--}\\
    \midrule
    ViT-B/16 (Ours) & {Key} & 4 & {Full} & \bf 83.40 & 78.76 & \bf 81.45 & \bf 81.89 & \bf 82.17\\
    ViT-B/16 (Ours) & {Key} & 8 & {Full} & 82.62 & 78.84 & 80.68 & 81.60 & 81.46\\
    ViT-B/16 (Ours) & {Key} & 16 & {Full} & 82.02 & 78.69 & 79.58 & 81.10 & 80.67\\
    \midrule
    ViT-B/16 (Ours) & {Key} & 4 & {LoRA ($r = 16$)} & 81.12 & 78.44 & 51.56 & 80.04 & 53.26\\
    ViT-B/16 (Ours)& {Key} & 8 & {LoRA ($r = 16$)} & 79.94 & 78.69 & 7.62 & 79.94 & 7.63\\
    ViT-B/16 (Ours) & {Key} & 16 & {LoRA ($r = 16$)} & 79.71 & \bf 78.96 & 0.73 & 79.42 & 0.72\\
    \midrule
    ConvMixer-768/32 (\cite{maung2023hindering}) & {Key} & 7 & {Partial} & 71.98 & 64.74 & 70.65 & 70.16 & 71.47\\
    ResNet50 (\cite{aprilpyone2021block}) & {Key} & 4 & {Full} & 75.69 & 66.95 & {--} & {--} & {--}\\
    Swin-L (\cite{liu2023comprehensive}) & {AT} & {--} & {--} & 78.92 & 59.56 & {--} & {--} & {--}\\
    ConvNeXt-L (\cite{liu2023comprehensive}) & {AT} & {--} & {--} & 78.02 & 58.48 & {--} & {--} & {--}\\
    ConvNeXt-L + ConvStem (\cite{singh2023revisiting}) & {AT} & {--} & {--} & 77.00 & 57.70 & {--} & {--} & {--}\\
    \bottomrule
  \end{tabular}
  }
\end{table*}

\subsection{Training Efficiency}
Training a model on the ImageNet-1k dataset from scratch requires a lot of resources and is not feasible for normal users with limited resources.
However, by using the latest fine-tuning techniques like~\cite{steiner2022how,hu2022lora}, one can train a key-based defended model in less than a day.
Table~\ref{tab:params} shows the trainable parameters of a ViT-B/16 model on full fine-tuning and LoRA fine-tuning.
LoRA is no doubt efficient in fine-tuning because it updates only a fraction of total trainable parameters.
In our experiments, we applied LoRA only on the patch embedding layer and qkv layers of all transformer blocks in the ViT-B/16.
We utilized 4 NVIDIA A100 GPU cards on the full fine-tuning, which took approximately 900 minutes.
LoRA training took about 1,200 minutes on one A100 GPU only.
Although LoRA is significantly efficient, adaptive attacks defeated LoRA fine-tuned models in our experiments.
We shall improve LoRA fine-tuned models to be more robust against adversarial examples in our future work.

\begin{table}
  \caption{Trainable Parameters of ViT/B16 with or without LoRA\label{tab:params}}
  \centering
  \begin{tabular}{lcc}
  \toprule
  {Model} & \multicolumn{2}{c}{\# Trainable Parameters ($\times 10^6$)}\\
          & {Full} & {LoRA} \\
  \midrule
  ViT-B/16 & 86.57 & \bf 1.38 (1.57\%)\\
  \bottomrule
  \end{tabular}
\end{table}

\section{Discussion\label{sec:discussion}}
\noindent{\bf Applicability.}
Without changing the model architecture, the proposed key-based defense can easily train many different models that yield a similar performance from a pre-trained model by varying keys.
Although we adopt block-wise pixel shuffling as the key-based transformation in this paper, other key-based transformations can be applied.
We demonstrate that we could train several key-based models even on ImageNet with a limited amount of resources without severely degrading the performance accuracy.
With recent hardware/platform availability, the one-key-one-model image classification paradigm is feasible and has the potential to defend against adversarial examples in real-world settings (\eg, self-driving cars).
Moreover, the previous works also showed that key-based models are diverse and can be used in an ensemble for traditional image classification scenarios~\cite{taran2020machine,maungmaung2021ensemble}.

\vspace{2mm}\noindent{\bf Limitations.}
In this paper, we deployed p-norm bounded adversaries (specifically AA strategy~\cite{croce2020reliable}), which is also used in RobustBench~\cite{croce2020robustbench}.
However, in real-world settings, the adversary is unknown.
Another limitation is that we applied a LoRA dimension, $r = 16$, and a value of scaling factor $16$ only.
It is our first attempt to use LoRA, and we showed that key-based models can easily be obtained by using the latest fine-tuning techniques, and it is feasible to deploy key-based models in one-key-one-model application settings.
Further investigation on LoRA fine-tuning on key-based models for adversarial robustness is required, and we shall pursue this direction in our future work.

\vspace{2mm}\noindent{\bf Future Work.}
We shall further investigate different attacks, such as patch attacks and unrestricted adversarial examples, to evaluate key-based models.
We shall also improve adversarial robustness in LoRA fine-tuned models in our future work.

\section{Conclusion\label{sec:conclusion}}
In this paper, we proposed to leverage pre-trained models and use the latest fine-tuning techniques to a key-based defense so that key-based defended models can be easily proliferated even on the ImageNet scale.
We stress that such defended models can potentially be deployed on devices for one-key-one-model application scenarios.
Our proposed fine-tuned models can easily be obtained even on limited computing resources.
Experiment results showed that our fine-tuned models achieved a comparable clean accuracy compared to a non-defended model.
Moreover, given the condition that attackers do not know the secret key, our fine-tuned models outperformed state-of-the-art models in terms of both clean and robust accuracy.
We also carried out adaptive attacks to further evaluate the proposed fine-tuned models.
The results confirmed that our fully fine-tuned models are resistant to adaptive attacks, and LoRA fine-tuned models, in their current form, are not robust against adaptive attacks.

\bibliographystyle{IEEEtran}
\bibliography{IEEEabrv,ref}

\end{document}